\documentclass{article}

\usepackage{times}
\usepackage{graphicx}
\usepackage{subfigure} 

\usepackage{natbib}

\usepackage{algorithm}
\usepackage{algorithmic}

\usepackage{hyperref}

\usepackage{amsmath}
\usepackage{amsfonts}
\usepackage{amssymb}

\newcommand{\fig}[1]{Figure~\ref{#1}}
\newcommand{\kl}[2]{\mathit{KL}\big(#1 || #2\big)}
\newcommand{\klit}{\mathit{KL}}
\newcommand{\logit}{\mathrm{logit}}
\newcommand{\supplementgate}[2]{#1 #2}

\usepackage[accepted]{arxiv}

\begin{document} 

\supplementgate{
\twocolumn[
\icmltitle{Comparison of Maximum Likelihood and GAN-based training of Real NVPs}



\icmlsetsymbol{equal}{*}

\begin{icmlauthorlist}
\icmlauthor{Ivo Danihelka}{dm,complex}
\icmlauthor{Balaji Lakshminarayanan}{dm}
\icmlauthor{Benigno Uria}{dm}
\icmlauthor{Daan Wierstra}{dm}
\icmlauthor{Peter Dayan}{gatsby}
\end{icmlauthorlist}

\icmlaffiliation{dm}{Google DeepMind, London, United Kingdom}
\icmlaffiliation{complex}{CoMPLEX, Computer Science, UCL}
\icmlaffiliation{gatsby}{Gatsby Unit, UCL}

\icmlcorrespondingauthor{Ivo Danihelka}{danihelka@google.com}

\icmlkeywords{GAN}

\vskip 0.3in
]



\printAffiliationsAndNotice{}  

\begin{abstract}
We train a generator by maximum likelihood
and we also train the same generator architecture by Wasserstein GAN.
We then compare the generated samples, exact log-probability densities
and approximate Wasserstein distances.
We show that an independent critic trained to approximate Wasserstein distance between the validation set and the generator distribution helps detect overfitting. 
Finally, we use ideas from the one-shot learning literature to develop a novel fast learning critic.
\end{abstract}

\section{Introduction}
There has been a substantial recent interest in deep generative models.  Broadly speaking, generative models can be classified into \emph{explicit} models where we have access to the model likelihood function, and \emph{implicit} models which provide a sampling mechanism for generating data, but do not need to explicitly define a likelihood function. Examples of explicit models are variational auto-encoders (VAEs) \citep{kingma2013auto, rezende2014stochastic} and PixelCNN \citep{oord2016conditional}. Examples of implicit generative models are generative adversarial networks (GANs) \citep{goodfellow2014generative} and stochastic simulator models. Explicit models are typically trained by maximizing the likelihood or its lower bound; they are popular in probabilistic modeling as the training procedure optimizes a well-defined quantity and the likelihood can be used for model comparison and selection. However likelihood is not about human perception \cite{theis2015note}.

Recently, GAN-based training has emerged as a promising approach for learning implicit generative models. In particular, generators trained using GAN-based approaches  
have shown to be capable of generating  photo-realistic samples.
There is a lot of interest in understanding the 
the properties of the learned generators in GANs and comparing them with traditional likelihood-based methods. One step in this direction was the quantitative analysis by  \citet{wu2016quantitative} who 
evaluate the (approximate) likelihood of decoder based models using 
annealed importance sampling (AIS).  We use a different approach to provide additional insights. We use a tractable generator architecture for which the log-probability densities can be computed exactly. We train the generator architecture by maximum likelihood 
and we also train the \textit{same} generator architecture by GAN. We then compare the properties of the learned generators.

\begin{figure}[tb]
\vskip 0.1in
\begin{center}
\centerline{\includegraphics[width=0.36\columnwidth]{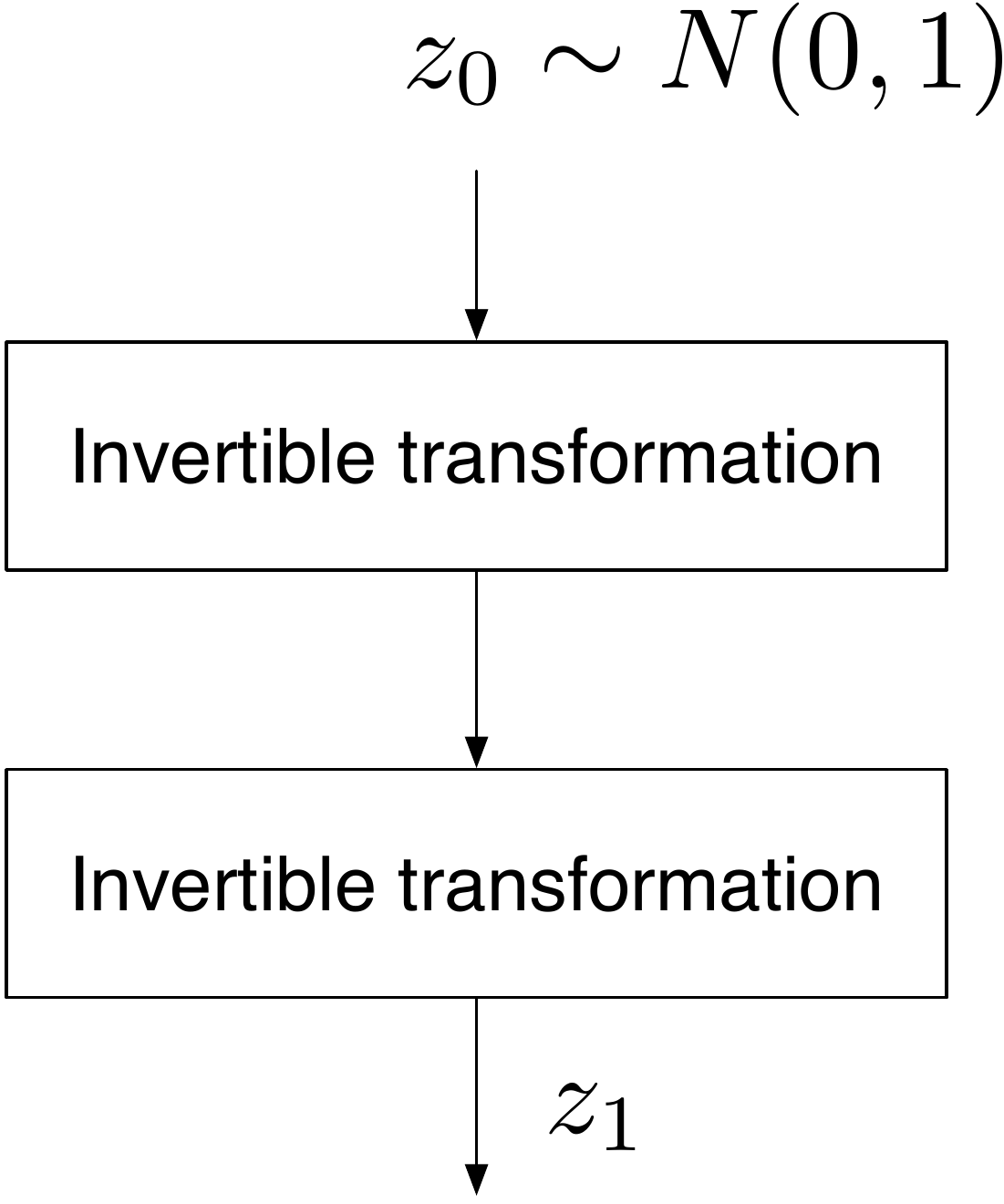}}
\caption{The generator is a real NVP with multiple invertible transformations.}
\label{nvp_decoder}
\end{center}
\vskip -0.26in
\end{figure} 

Concretely, we use real non-volume preserving transformation (real NVP) \citep{dinh2016density} as the generator. We compare training by maximum likelihood to training by Wasserstein GAN (WGAN)  \citep{arjovsky2017wasserstein}. In WGAN, a critic learns to approximate the Wasserstein distance between the data and the generator distribution. The generator 
learns to minimize the critic's approximation of the  Wasserstein distance.
WGAN has a well defined training procedure and allowed us to train a non-traditional GAN generator.
Another advantage of WGAN is that the critic's approximation of the Wasserstein distance can be used to detect overfitting as well as to compare different models. We discuss this more in Section~\ref{sec:wdistance}.


\begin{figure*}[tb]
\vskip 0.2in
\begin{center}
\begin{minipage}{0.46\textwidth}
\centerline{\includegraphics[width=\columnwidth]{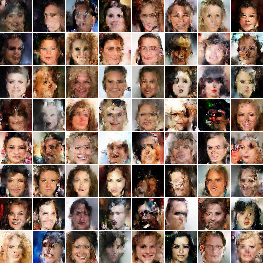}}
\end{minipage}
\hskip 0.2cm
\begin{minipage}{0.46\textwidth}
\centerline{\includegraphics[width=\columnwidth]{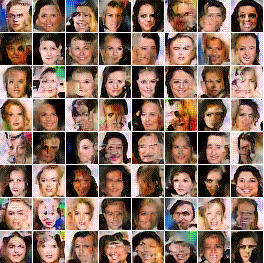}}
\end{minipage}
\caption{\textbf{Left:} Samples from a generator NVP3 trained by maximum likelihood. \textbf{Right:} Samples from a generator NVP3 trained by WGAN. Both generators had the same architecture. The WGAN samples would look better if using the improved training of WGAN-GP \citep{gulrajani2017improved}. The other conclusions are similar for WGAN-GP.}
\label{fig:nvp_333}
\end{center}
\vskip -0.2in
\end{figure*}
\begin{figure*}[tb]
\vskip 0.2in
\begin{center}
\begin{minipage}{0.46\textwidth}
\centerline{\includegraphics[width=\columnwidth]{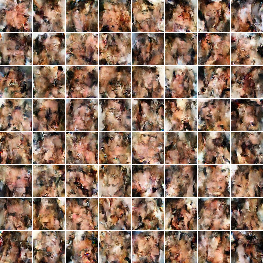}}
\end{minipage}
\hskip 0.2cm
\begin{minipage}{0.46\textwidth}
\centerline{\includegraphics[width=\columnwidth]{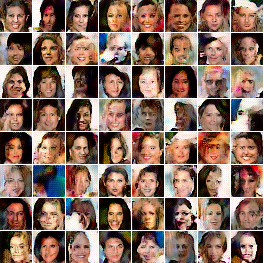}}
\end{minipage}
\caption{\textbf{Left:} Samples from the shallow generator NVP1 trained by maximum likelihood. \textbf{Right:} Samples from the same shallow architecture trained by WGAN.}
\label{fig:nvp_3}
\end{center}
\vskip -0.2in
\end{figure*} 

WGAN worked for us better than other GAN methods (Jensen-Shannon divergence or $-\log D(x)$).
We intend to communicate the found interesting differences from training by maximum likelihood.
After describing the generator and the critic in Section~\ref{sec:setup},
we report some perhaps surprising
findings in Section~\ref{sec:results}: 
\begin{enumerate}
\item Generators trained by WGAN produce more globally coherent samples even from a relatively shallow generator.
\item Minimization of the approximate Wasserstein distance does not correspond to minimization of the negative log-probability density. The negative log-probability densities became worse than densities from a uniform distribution.
\end{enumerate}

We also report a few findings about the approximate Wasserstein distance:
\begin{enumerate}
\item An approximation of the Wasserstein distance ranked correctly generators trained by maximum likelihood.
\item The approximate Wasserstein distance between the training data and the generator distribution became smaller than the distance between the test data and the generator distribution. This overfitting was observed for generators trained by maximum likelihood and also for generators trained by WGAN.
\end{enumerate}

Finally, we use the ability to compare different models
when evaluating a novel fast learning critic in Section~\ref{sec:fast_critic}.
 
\section{Setup}
\label{sec:setup}
\subsection{Generator}
\label{sec:generator}

In all our experiments, the generator is a real-valued non-volume preserving transformation (NVP) \citep{dinh2016density}. The generator is schematically represented in \fig{nvp_decoder}. The generator starts with \textit{latent} noise $z_0$ from the standard normal distribution. The noise is transformed by a sequence of invertible transformations to a generated image $z_1$.
Like other generators used in GANs, the generator does not add independent noise to the generated image.

Unlike other GAN generators, we are able to compute the log-probability density of the generated image $z_1$ by:
\begin{align}
\label{eq:logpz}
   \log p_{Z_1}(z_1) = \log p_{Z_0}(z_0) - \log |\det \frac{\partial z_1}{\partial z_0} | 
\end{align}
where $\log p_{Z_0}(z_0)$ is the log-probability density of the standard normal distribution. The determinant of the $\frac{\partial z_1}{\partial z_0}$ Jacobian matrix can be computed efficiently, if each transformation has a triangular Jacobian \citep{dinh2016density}.

We can also compute the log-probability density assigned to an image
by running the inverse of the transformations. The inferred $z_0$ can be then plugged to Equation~\ref{eq:logpz}.

We use a generator with the convolutional multi-scale architecture designed by \citet{dinh2016density}.
Concretely, we use 3 different generator architectures with 1, 2 and 3 multi-scale levels. In figures, these generators are labeled NVP1, NVP2, NVP3 and have 7, 13 and 19 non-volume preserving transformations. Each transformation uses 4 residual blocks. In total, the generator NVP3 has $19 \times 4 = 76$ layers. It is a very deep generator. To have the same training and testing conditions, we use no batch normalization in the generator.

\subsection{Critic}
Wasserstein GAN (WGAN) \citep{arjovsky2017wasserstein} uses a critic instead of a discriminator.
The critic is trained to provide an approximation of the Wasserstein distance between the real data distribution $P_r$ and the generator distribution $P_g$. The approximation is based on the Kantorovich-Rubinstein duality:
\begin{align}
\label{eq:duality}
W(P_\mathit{r}, P_\mathit{g}) \propto \max_f \mathbb{E}_{x \sim P_\mathit{r}}\left[f(x)\right] - \mathbb{E}_{x \sim P_\mathit{g}}\left[f(x)\right]
\end{align}
where $f: \mathbb{R}^D \to \mathbb{R}$ is a Lipschitz continuous function. In practice, the critic $f$ is a neural network with clipped weights to have bounded derivatives. The critic is trained to produce high values at real samples and low values at generated samples. The difference of the expected critic values then approximates the Wasserstein distance.
The approximation is scaled by the Lipschitz constant for the critic \citep{arjovsky2017wasserstein}.
After training the critic for the current generator distribution,
the generator can be trained to minimize the approximate Wasserstein distance.

We follow closely the original WGAN implementation\footnote{\url{https://github.com/martinarjovsky/WassersteinGAN}} and use a DCGAN \citep{radford2015unsupervised} based critic. The implementation uses learning rate 0.00005, batch size 64
and clips the critic weights to a fixed box $[-0.01, 0.01]$. The critic is updated 100 times in the first 25 generator steps and also after every 500 generator steps, otherwise the critic is updated 5 times per generator step.

We rescale the images to $[-1, 1]$ range before feeding them to the critic.

\subsection{Datasets}
We train on $28 \times 28$ images from MNIST \citep{lecun1998gradient} and on $32 \times 32$ images from the CelebFaces Attributes Dataset (CelebA) \citep{liu2015faceattributes}. The CelebA dataset contains over 200,000 celebrity images.
During training we augment the CelebA dataset to also include horizontal flips of the training examples
as done by \citet{dinh2016density}.

\subsection{Data Preprocessing}
We convert the discrete pixel intensities from $\{0, \dots, 255\}^D$ to continuous noisy images from $[0, 256]^D$ by adding a uniform noise $u \in [0, 1]^D$ to each pixel intensity. The computed log-probability density will be then a lower bound for the log-probability mass \cite{theis2015note}:
\begin{align}
   \int_{[0, 1]^D} \log p(x + u)\,du \leq \log \int_{[0, 1]^D} p(x + u)\,du
\end{align}

Finally, before feeding the pixel intensities to a network,
we divide the pixel intensities by 256. This is a simple non-volume preserving transformation.
If the noisy image is $z_2 = x + u$ and the image with scaled intensities is $z_1 = \frac{z_2}{256}$ then the log-probability density of the noisy image is:
\begin{align}
   \log p_{Z_2}(z_2) &= \log p_{Z_1}(z_1) - \log |\det \frac{\partial z_2}{\partial z_1} | \\
     &= \log p_{Z_1}(z_1) - D\log 256
\end{align}
where $D$ is the number of dimensions in the image (e.g., $D = 32 \times 32 \times 3$).
For example, if an uninformed model assigns uniform probability density $p_{Z_1}(z_1)=1$ to all $z_1 \in [0, 1]^D$,
the model would require $\log_2(256) = 8$ bits/dim to describe an image.
The reported negative log-probability density then corresponds to an upper bound on compression loss in bits/dim.

\section{Results}
\label{sec:results}
 
We trained the same generator architecture by maximum likelihood
and by WGAN. We will now compare the effect of these two different objectives.

\subsection{Generated Samples}
\fig{fig:nvp_333} shows samples from a generator trained by maximum likelihood and from another generator trained by WGAN. Both generators have the same number of parameters and the same architecture.

The generator trained by WGAN seems to produce more coherent faces.
The effect is more apparent when looking at samples produced from a shallower generator in \fig{fig:nvp_3}.
The shallow generator NVP1 learned to produce only locally coherent image patches if trained by maximum likelihood.

\begin{figure}[tb]
\vskip 0.2in
\begin{center}
\centerline{\includegraphics[width=0.8\columnwidth]{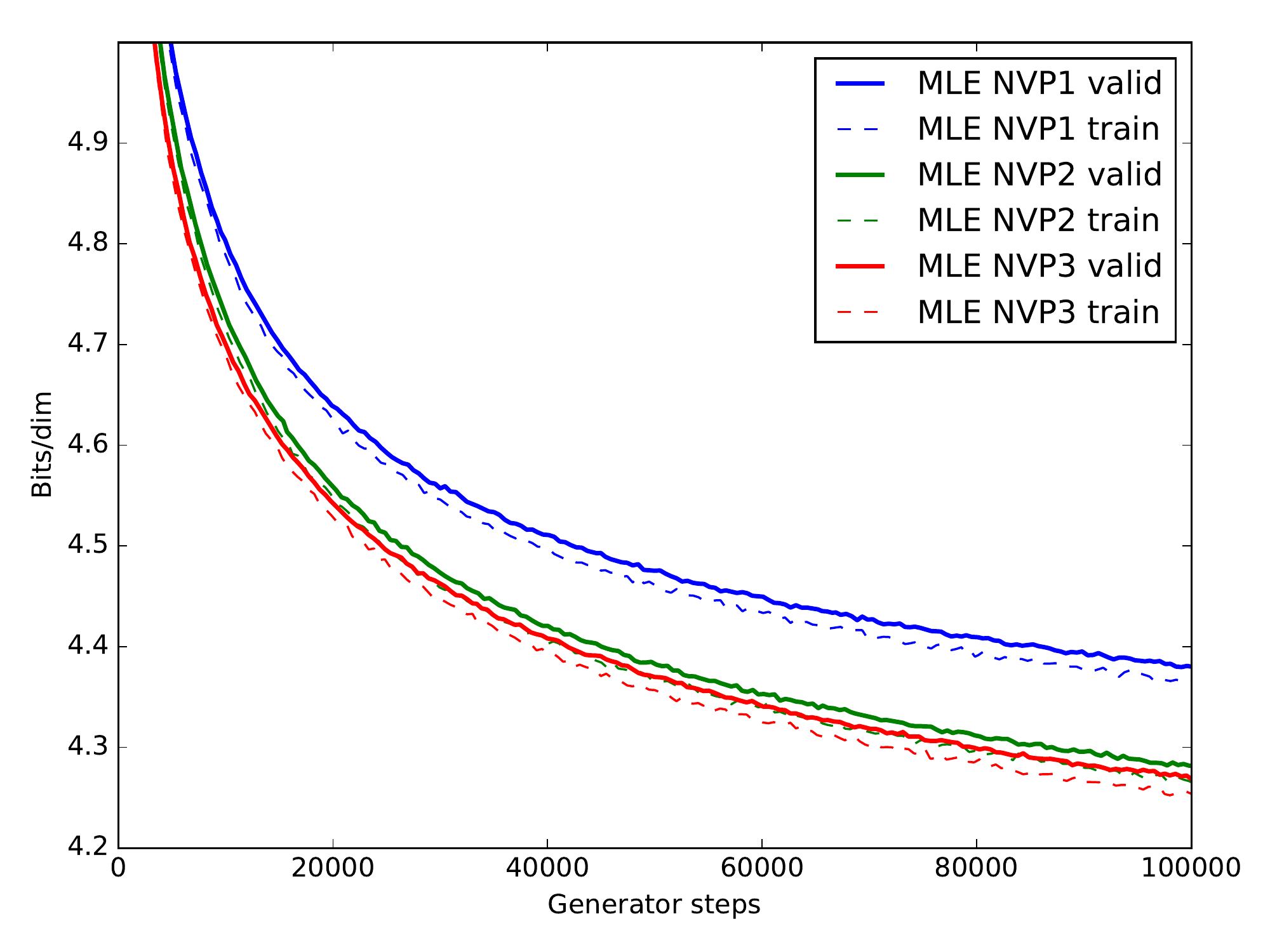}}
\caption{The negative log-probability density of real NVPs in \textit{bis/dim} on CelebA $32\times32$ images. The generators
were trained by maximum likelihood estimation (MLE). The loss is slightly higher on the validation set.}
\label{fig:plot_obs_bits_mle}
\end{center}
\vskip -0.2in
\end{figure} 

\begin{figure}[tb]
\vskip 0.2in
\begin{center}
\centerline{\includegraphics[width=0.8\columnwidth]{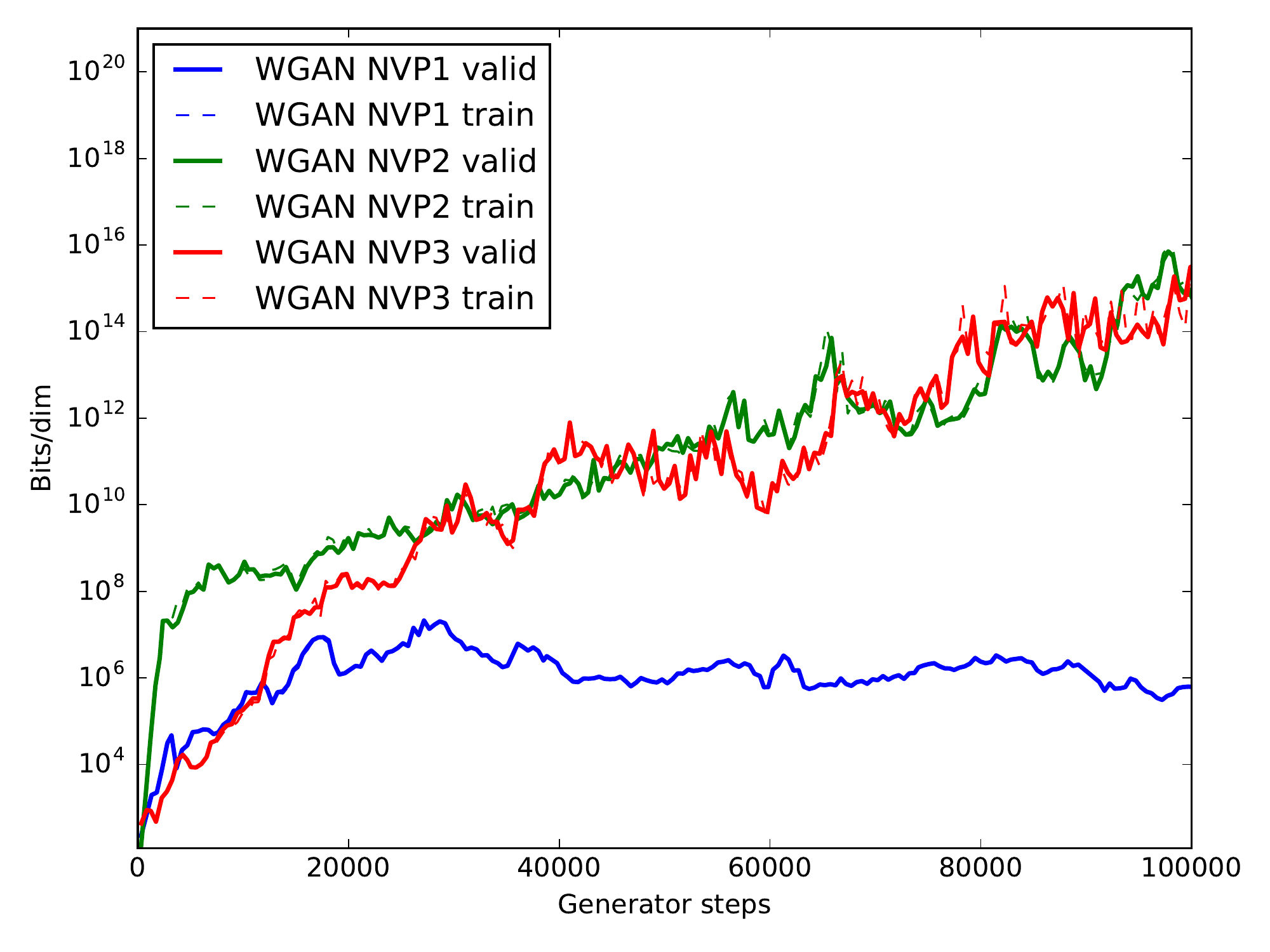}}
\caption{The negative log-probability density of real NVPs trained by WGAN on CelebA. Note the logarithmic scale on y-axis.}
\label{fig:plot_obs_bits_wgan}
\end{center}
\vskip -0.2in
\end{figure} 

\begin{figure}[tb]
\vskip 0.2in
\begin{center}
\centerline{\includegraphics[width=0.5\columnwidth]{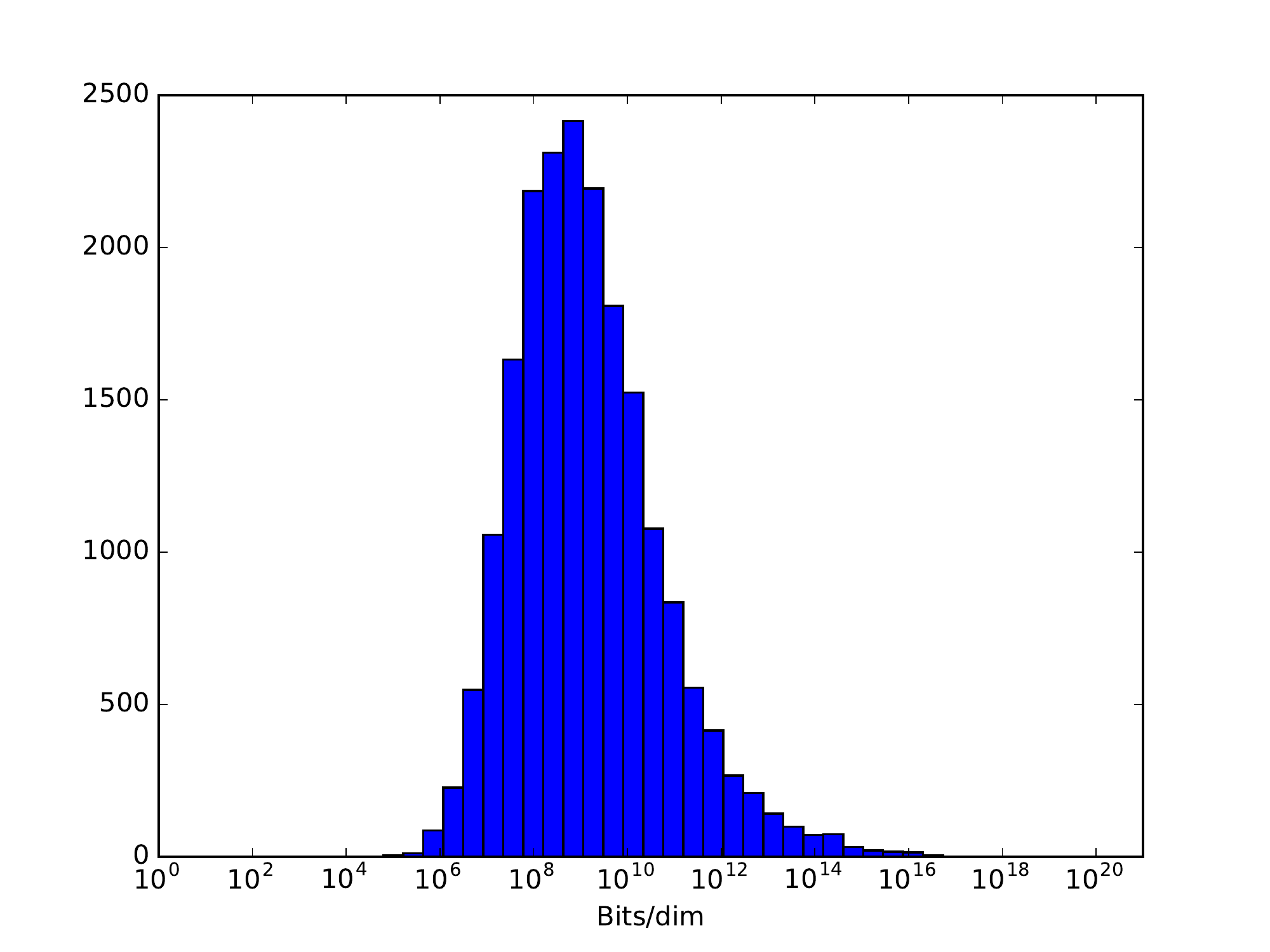}}
\caption{The histogram of negative log-probability densities on the CelebA validation set. The generator NVP3 was trained by WGAN.}
\label{fig:eval_histogram_of_logp_noreset2_nvp3_wgan}
\end{center}
\vskip -0.2in
\end{figure} 

\begin{figure}[tb]
\vskip 0.2in
\begin{center}
\begin{minipage}{0.23\textwidth}
\centerline{\includegraphics[width=\columnwidth]{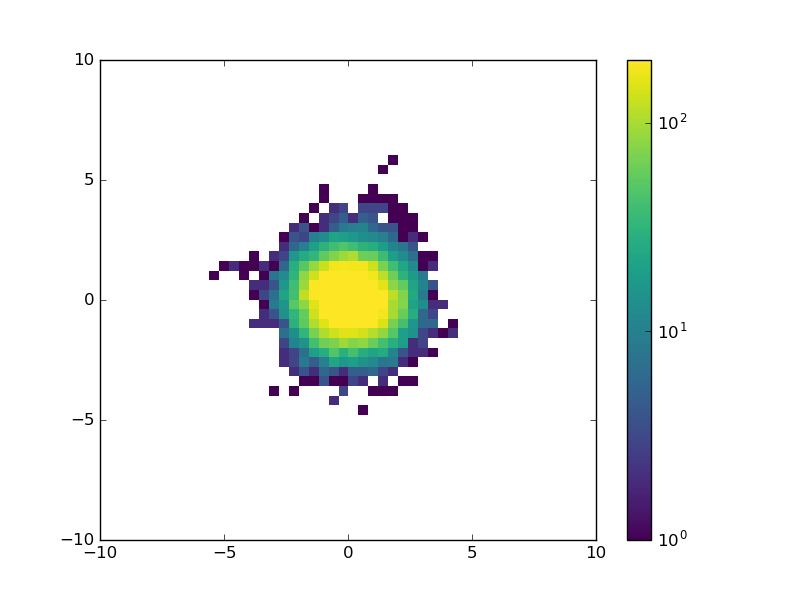}}
\end{minipage}
\begin{minipage}{0.23\textwidth}
\centerline{\includegraphics[width=\columnwidth]{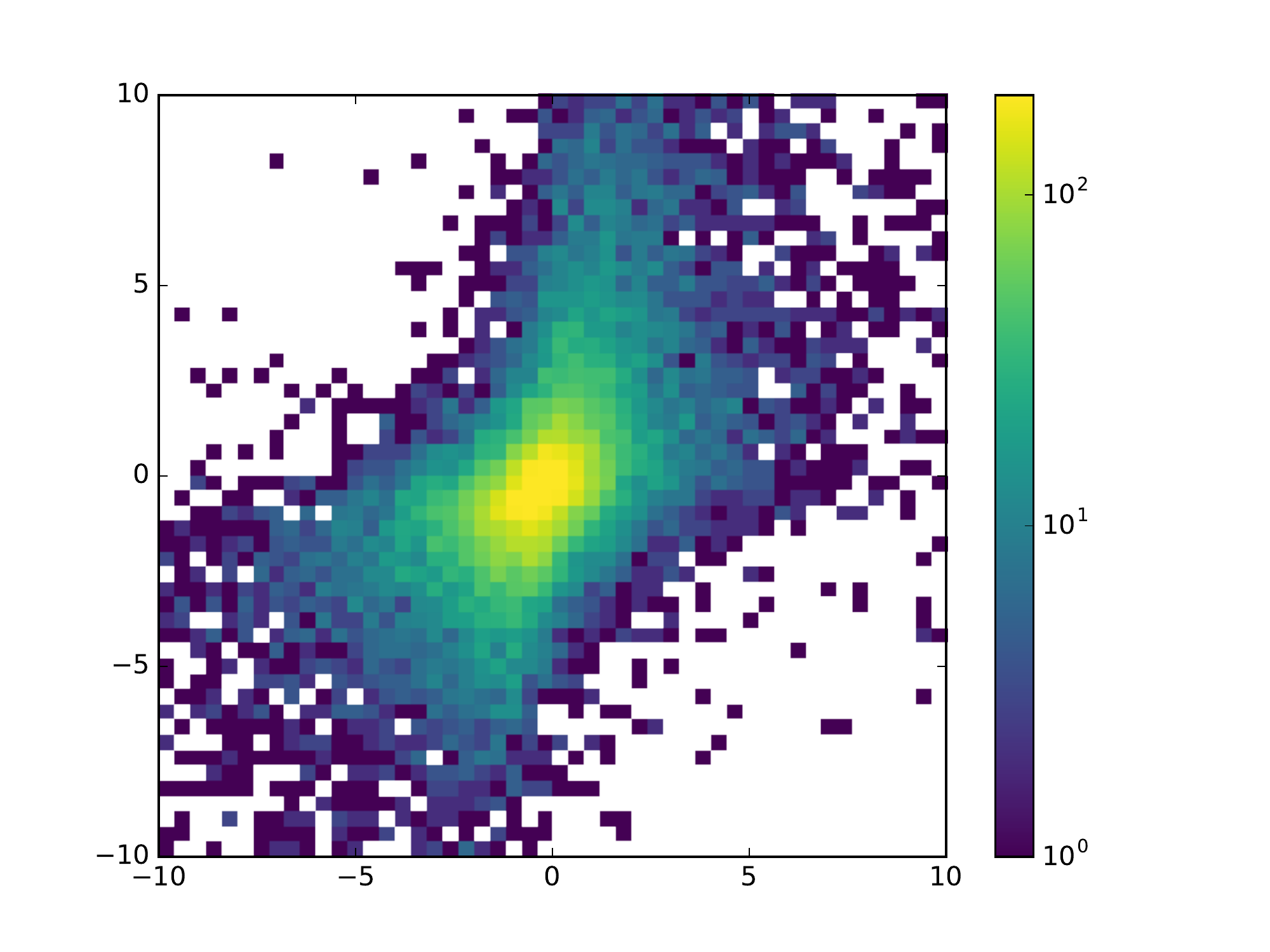}}
\end{minipage}
\caption{\textbf{Left:} 2D histogram of 2 latent variables for examples from the CelebA validation set. The generator was NVP3 trained by maximum likelihood. \textbf{Right:} 2D histogram of 2 latent variables for generator NVP3 trained by WGAN.}
\label{fig:eval_histogram_of_z}
\end{center}
\vskip -0.2in
\end{figure}

\begin{figure}[tb]
\vskip 0.2in
\begin{center}
\centerline{\includegraphics[width=\columnwidth]{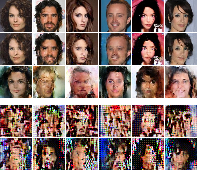}}
\caption{\textbf{From top to bottom:} \textbf{1)} Original images from the CelebA validation set. \textbf{2)} Reconstructions from the first half of latent variables. The generator was NVP3 trained by MLE. \textbf{3)} Reconstructions from the second half of latent variables for MLE. \textbf{4)} Reconstructions from the first half of latent variables if the generator was trained by WGAN. \textbf{5)} Reconstructions from the second half of latent variables for WGAN.}
\label{fig:reconstruction}
\end{center}
\vskip -0.2in
\end{figure} 

\begin{figure*}[tb]
\vskip 0.2in
\begin{center}
\begin{minipage}{0.46\textwidth}
\centerline{\includegraphics[width=\columnwidth]{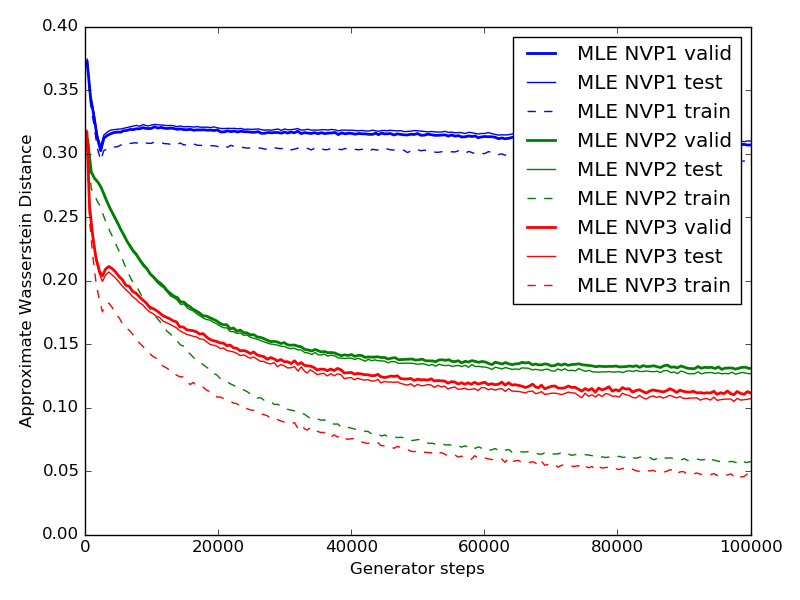}}
\end{minipage}
\begin{minipage}{0.46\textwidth}
\centerline{\includegraphics[width=\columnwidth]{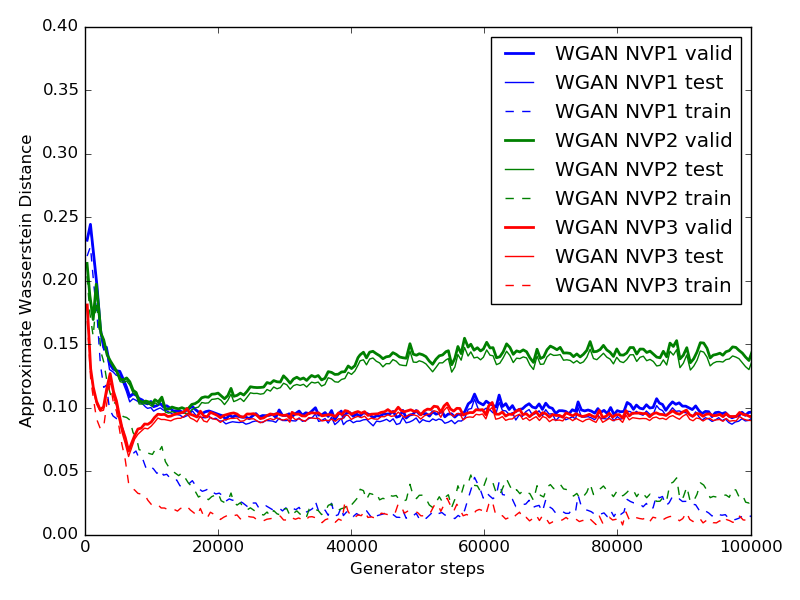}}
\end{minipage}
\caption{\textbf{Left:} The approximate Wasserstein distance for real NVPs trained by maximum likelihood on CelebA. \textbf{Right:} The approximate Wasserstein distance for real NVPs trained by WGAN. The Wasserstein distance was approximated by an independent critic.}
\label{fig:plot_wdistance}
\end{center}
\vskip -0.2in
\end{figure*}

\subsection{Log-probability Density}
\fig{fig:plot_obs_bits_mle} shows the negative log-probability density for generators trained by maximum likelihood estimation (MLE).
The shallowest generator NVP1 obtained the worse performance there. The generators show only a small amount of overfitting to the training set. The validation loss is slightly higher than the training loss.

Generators trained by WGAN have their negative log-probability densities shown in \fig{fig:plot_obs_bits_wgan}.
The figure has a completely different y-axis. The negative log-probability densities get worse when training by WGAN.
The training and validation losses overlap.

A generator can get infinite negative log-probability if it assigns zero probability to an example.
To check this possibility, we analysed the negative log-probabilities assigned to all validation examples.
\fig{fig:eval_histogram_of_logp_noreset2_nvp3_wgan} shows the histogram of negative log-probabilities. We see that all 19867 validation examples have low log-probabilities. A similar histogram was obtained also for the training set. So the bad average log-probability is not caused by a single outlier.

We further inspected the negative log-probability density of generated samples. We observed that the average negative log-probability density of generated samples is \textbf{-1.12} bits/dim. The negative sign is not a typo. The negative log-probability density can be negative, if the probability density is bigger than 1.
In contrast, the NVP3 generator trained by maximum likelihood assigned on average 4.27 bits/dim to the validation examples and 4.35 bits/dim to its own generated samples. 

The problematic negative log-probability densities from WGAN were not limited to the CelebA dataset. Generators trained by WGAN had high negative log-probability densities also on MNIST (\fig{fig:plot_mnist_obs_bits_wgan}).
A deep generator trained by WGAN learns a distribution lying on a low dimensional manifold (Figure~\ref{fig:plot_sval} in the appendix). The generator is then putting the probability mass only to a space with a near-zero volume. We may need a more powerful critic to recognize the excessively correlated pixels.
Approximating the likelihood by annealed importance sampling \citep{wu2016quantitative} would not discover this problem, as their analysis assumes a Gaussian observation model with a fixed variance.

The problem is not unique to WGAN. We also obtained near-infinite negative log-probability densities when training GAN to minimize the Jensen-Shannon divergence \citep{goodfellow2014generative}.

\subsection{Distribution of Latent Variables}
\label{sec:latent}
One of the advantages of real NVPs is that we can infer the original latent $z_0$ for a given generated sample. We know that the distribution of the latent variables is the prior $N(0, 1)$,
if the given images are from the generator.
We are curious to see the distribution of the latent variables, if the given images are from the validation set.

In \fig{fig:eval_histogram_of_z}, we display a 2D histogram of the first 2 latent variables $z_0[1], z_0[2]$. The histogram was obtained by inferring the latent variables for all examples from the validation set. When the generator was trained by maximum likelihood, the inferred latent variables had the following means and standard deviations: $\mu_1=0.05, \mu_2=0.05, \sigma_1=1.06, \sigma_2=1.03$.
In contrast, the generator trained by WGAN had inferred latent variables with significantly larger standard deviations: $\mu_1=0.02, \mu_2=1.62, \sigma_1=3.95, \sigma_2=8.96$.

When generating the latent variables from the $N(0, 1)$ prior, the samples from the generator trained by WGAN
would have a different distribution than the validation set.

\subsection{Partial Reconstructions}
Real NVPs are invertible transformations and have perfect reconstructions.
We can still visualize reconstructions from a partially resampled latent vector.
\citet{gregor2016towards} and \citet{dinh2016density} visualized `conceptual compression' by inferring the latent variables
and then resampling a part of the latent variables from the normal $N(0, 1)$ prior.
The subsequent reconstruction should still form a valid image.
If the original image was generated by the generator, the partially resampled latent vector would still have the normal $N(0, 1)$ distribution.

\fig{fig:reconstruction} shows the reconstructions if resampling the first half or the second half of the latent vector. The generator trained by maximum likelihood (MLE) has partial reconstructions similar to generated samples. In comparison, the partial reconstructions from the generator trained by WGAN do not resemble samples from WGAN. This again indicates that the validation examples have a different distribution than WGAN samples.

\subsection{Wasserstein Distance}
\label{sec:wdistance}
We will now look at the approximate Wasserstein distance between the validation data and the generator distribution. To approximate the Wasserstein distance, we will use the duality in Equation~\ref{eq:duality}. We will train another critic to assign high values to validation samples and low values to generated samples. This \textit{independent} critic will be used only for evaluation. The generator will not see the gradients from the independent critic.

The independent critic is trained to maximize:
\begin{align}
\label{eq:indep_critic}
  \hat{W}(x_\mathit{valid}, x_g) &= \frac{1}{N}\sum^{N}_{i=1} \hat{f}(x_\mathit{valid}[i]) - \frac{1}{N}\sum^{N}_{i=1} \hat{f}(x_g[i]))
\end{align}
where $x_\mathit{valid}$ is a batch of samples from the validation set, $x_g$ is a batch of generated samples and $\hat{f}$ is the independent critic.

We will keep the Lipschitz constant of the independent critic approximately constant by always using same architecture and the same weight clipping for the independent critic. We can then compare the approximate Wasserstein distances from different experiments.

\fig{fig:plot_wdistance}-left shows the approximate Wasserstein distance between the validation set and the generator distribution. The first thing to notice is the correct ordering of generators trained by maximum likelihood. The deepest generator NVP3 has the smallest approximate distance from the validation set, as indicated by the thick solid lines.

We also display an approximate distance between the training set and generator distribution:
\begin{align}
\hat{W}(x_\mathit{train}, x_g)
\end{align}
and the approximate distance between the test set and the generator distribution:
\begin{align}
\hat{W}(x_\mathit{test}, x_g)
\end{align}
where $x_\mathit{train}$ is a batch of training examples, $x_\mathit{test}$ is a batch of test examples and $\hat{W}$ is the approximate Wasserstein distance
computed by Equation~\ref{eq:indep_critic}.
We are misusing the independent critic here.
We are asking the independent critic to assign values to training  and test examples. The independent critic was trained only to assign values to validation examples and to generated samples.
This leads to a desirable effect: We can detect whether the generator overfits the training data.
If the training examples have the same distribution as the test examples, we should observe:
\begin{align}
 \mathbb{E}\left[\hat{W}(x_\mathit{train}, x_g)\right] = \mathbb{E}\left[\hat{W}(x_\mathit{test}, x_g)\right]
\end{align}
In practice, the empirical distribution of the training data is not the same as the distribution of the test data. The generated samples $x_g$ can be more similar to the training data
than to the test data.

\fig{fig:plot_wdistance} clearly shows that:
\begin{align}
\hat{W}(x_\mathit{train}, x_g) < \hat{W}(x_\mathit{test}, x_g)
\end{align}
for the trained generators. 
The approximate distance between the test set and the generator distribution
is a little bit smaller than the approximate distance between the validation set and the generator distribution.
The approximate distance between the training set and the generator distribution is much smaller. 
The generators are overfitting the training set.

In multiple experiments, we found that the performance of the generators is heavily influenced by the WGAN critic used for training. \fig{fig:plot_wdistance}-right shows that $\hat{W}(x_\mathit{valid}, x_g)$ is roughly the same for NVP1 and NVP3 (blue and red thick solid lines). Both generators were trained with the same WGAN critic architecture.
The generators can also overfit and gradually degrade the performance on the validation set. Visual inspection of the samples can be then misleading.

The approximate Wasserstein distance introduced by \citet{arjovsky2017wasserstein} is a very useful tool for model selection.
If we use the independent critic, we can compare generators trained by other GAN methods or by different approaches.
MNIST results are in \fig{fig:plot_mnist_wdistance}.
In the next section, we will use the independent critic to compare different critic architectures.

\begin{figure}[tb]
\vskip 0.2in
\begin{center}
\centerline{\includegraphics[width=0.3\columnwidth]{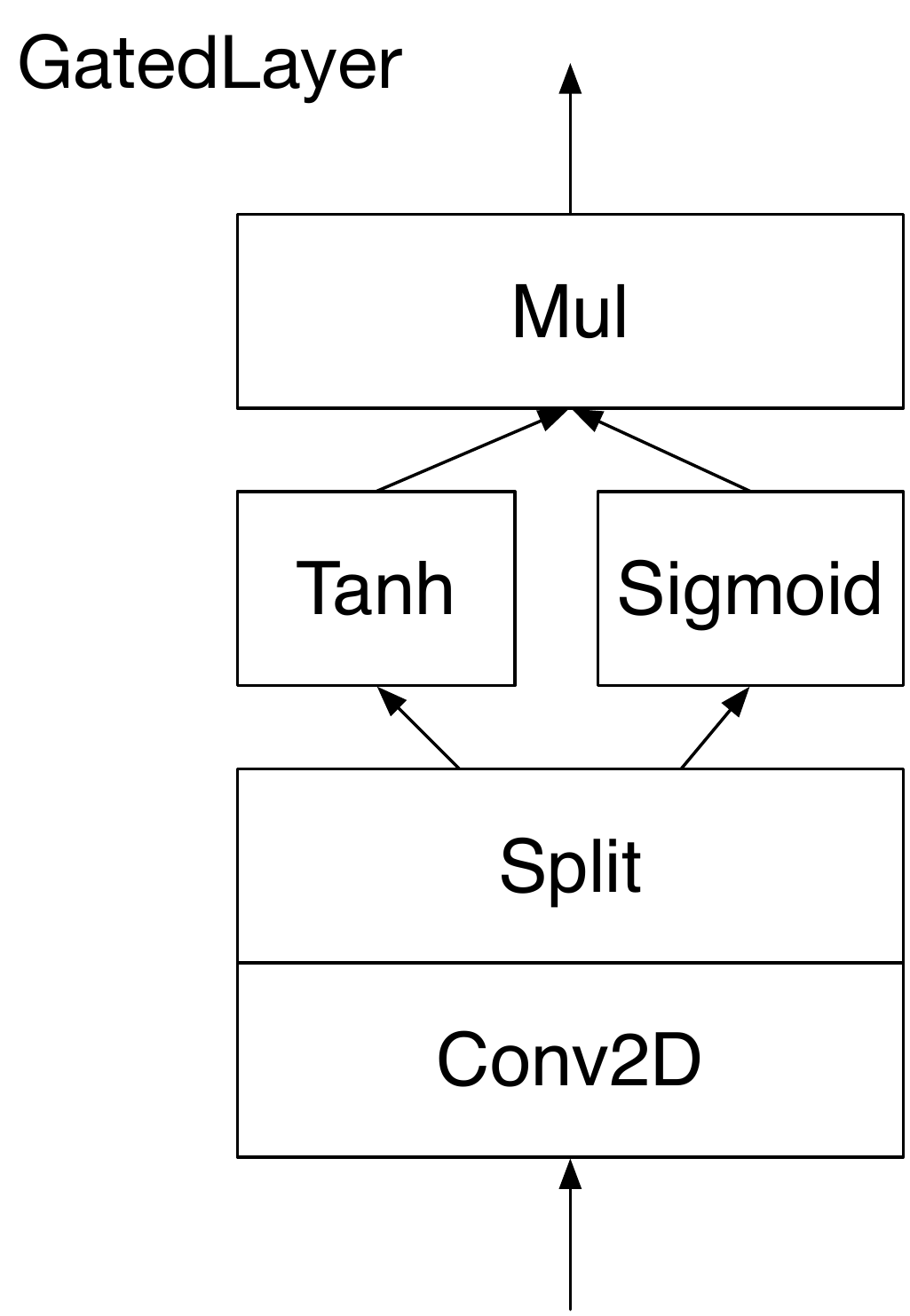}}
\caption{The gated convolutional layer from \citet{oord2016conditional}. The channels are split and passed to the element-wise multiplicative interaction.}
\label{fig:fast_critic_gated_conv}
\end{center}
\vskip -0.2in
\end{figure}

\begin{figure}[tb]
\vskip 0.2in
\begin{center}
\centerline{\includegraphics[width=\columnwidth]{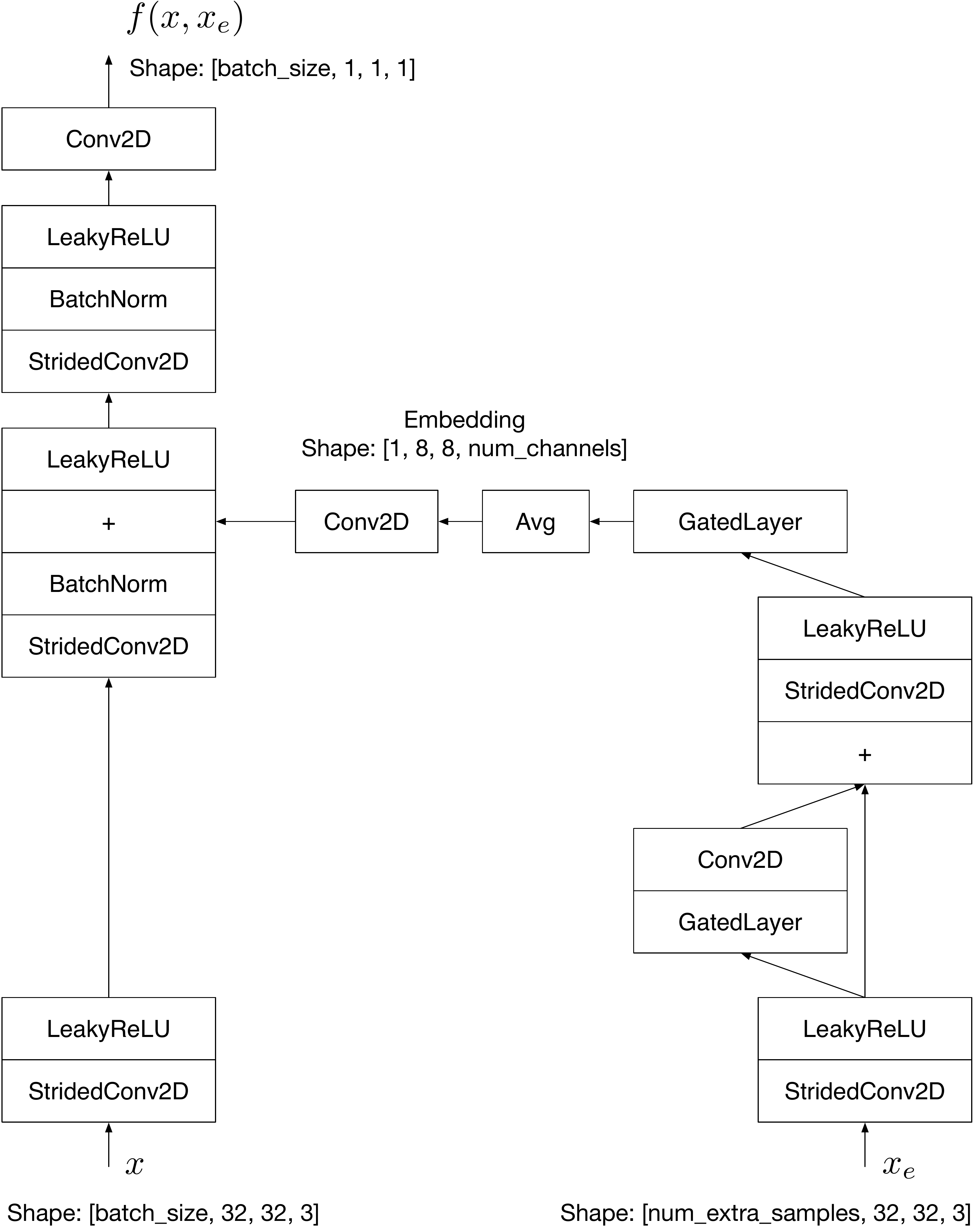}}
\caption{The fast learning critic $f(x, x_e)$ outputs the value for samples $x$,
conditioned on an extra batch of samples $x_e$ from the generator.}
\label{fig:fast_critic}
\end{center}
\vskip -0.3in
\end{figure}

\begin{figure}[tb]
\vskip 0.2in
\begin{center}
\centerline{\includegraphics[width=0.8\columnwidth]{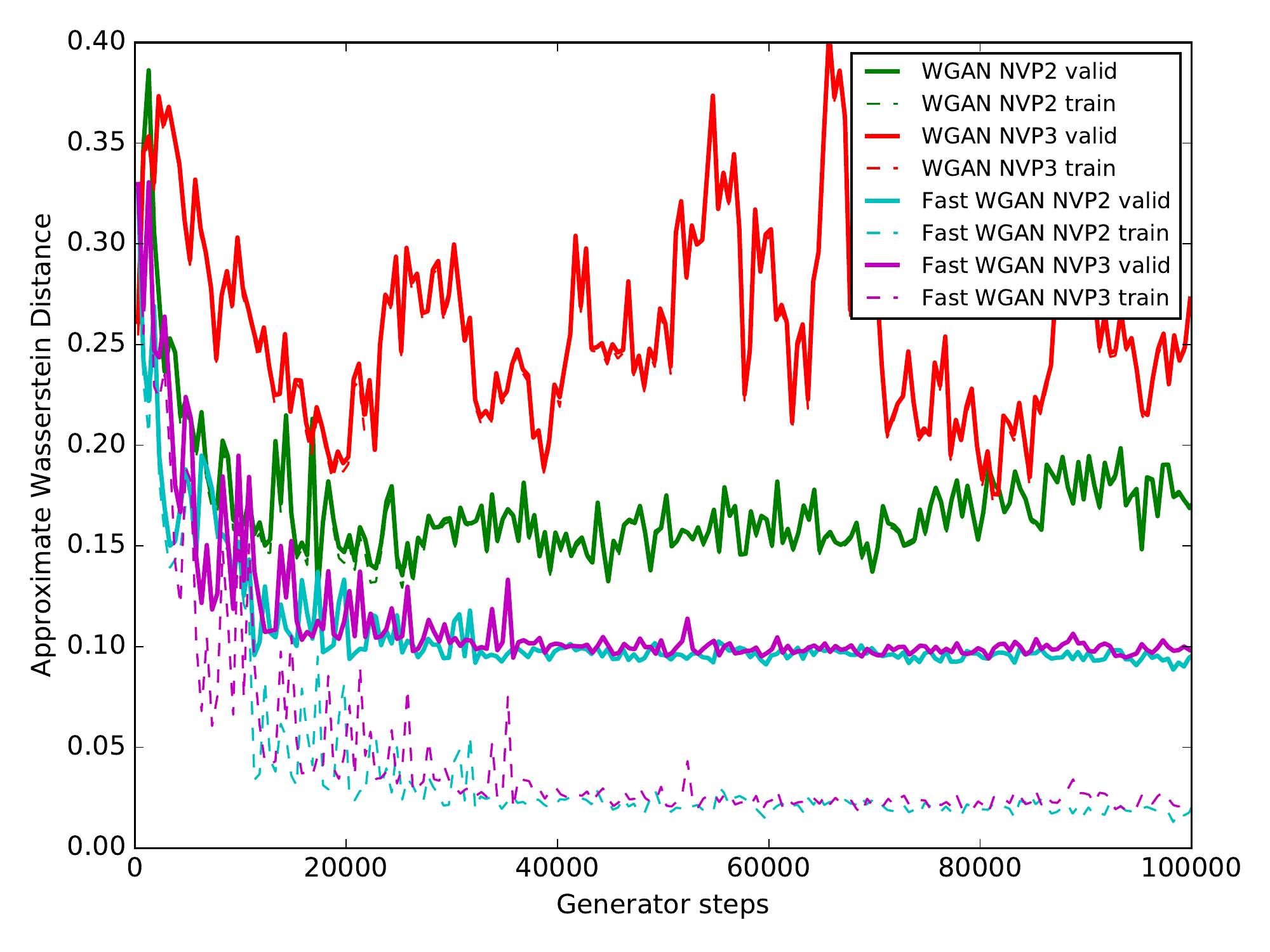}}
\caption{The approximate Wasserstein distance on CelebA when updating the critic less often. The shown Wasserstein distance was approximated by an independent critic.}
\label{fig:plot_wdistance_wgan_diters}
\end{center}
\vskip -0.3in
\end{figure}

\begin{algorithm}[t]
   \caption{Fast Learning Critic}
   \label{alg:fast_critic}
\begin{algorithmic}
   \STATE {\bfseries Require:} batch size $N$, conditional critic $f$.
   \STATE Sample $x_r \sim P_r$ a batch from the real data.
   \STATE Sample $x_g \sim P_g$ a batch of generator samples.
   \STATE Sample $x_\mathit{e} \sim P_g$ extra generator samples.
   \STATE Compute the approximate Wasserstein distance:
   \STATE $W := \frac{1}{N} \sum^N_{i=1} f(x_r[i], x_\mathit{e}) - \frac{1}{N} \sum^N_{i=1} f(x_g[i], x_\mathit{e})$
\end{algorithmic}
\end{algorithm}

\section{Fast Learning Critic}
\label{sec:fast_critic}
To obtain good results, WGAN requires to have a good critic.
The critic needs to be retrained when the generator changes.
It would be nice to reduce the number of needed retraining steps.
We will take ideas from supervised setting where people used a short-term memory to provide one-shot learning \citep{santoro2016one,vinyals2016matching}.

If we look at the optimization problem for the critic (Equation~\ref{eq:duality}),
we see that the optimal critic would need to be based
on the data distribution and on the \textit{current} generator distribution.
The data distribution is stationary,
so the knowledge of the data distribution can be kept stored
in the critic weights.
The generator distribution is changing quickly.
So we should give the critic a lot of information about the current
generator distribution.

For example, it would be helpful to generate extra samples from the generator
and to compute moments of the generator distribution.
The moments can be then passed as extra information to the critic.
We will do something more powerful. We will allow the critic
to extract features from the extra generator samples.

Concretely, we will allow the critic to condition on the extra samples from the generator. The extra samples are processed to produce a learned embedding of the generator distribution \citep{muandet2016kernel}. The embedding is then used to bias the critic.

\subsection{Architecture}
The implemented critic architecture is depicted in \fig{fig:fast_critic}.
The architecture looks like a DCGAN discriminator \citep{radford2015unsupervised} conditioned on an embedding.
The distribution embedding is produced by a network with gated activations (\fig{fig:fast_critic_gated_conv}) on the residual connections \citep{oord2016conditional}. The features from the batch of extra generator samples are averaged over the batch dimension
to produce the distribution embedding.

Algorithm~\ref{alg:fast_critic} shows the usage of the critic. 
The embedding remains the same when running the critic on real or generated samples.
The critic would become the original WGAN critic, if using zeros instead of the distribution embedding.
We use batch size 64 and we also use 64 extra generator samples.

The weights used to produce the embedding of the distribution do not need to be clipped and we do not clip them.

When training a generator, we do not use the gradients with respect to the extra generator samples.
The generator is only trained with the gradient of the approximate Wasserstein distance with respect to $x_g$.

\subsection{Fast Learning Results}
\fig{fig:plot_wdistance_wgan_diters} compares training without and with the fast learning critic. The critic was intentionally updated less frequently by gradient descent to demonstrate the benefits of the fast learning. The critic was updated 100 times in the first 25 generator steps and also after every 500 generator steps, otherwise the critic is updated only \textbf{2} times per generator step. The independent critic was still updated at least 5 times per generator step to keep all measurements comparable.

Without the fast learning critic, the generator failed to produce samples
similar to the data examples.
The fast learning critic may be important for conditional models
and for video modeling. We do not have multiple real samples for a situation there.

\section{Discussion}
In our experiments, we used two tools for the evaluation of generators.
First, we used a real NVP to compute the exact log-probability densities.
Second, we used an independent critic to compare the approximate Wasserstein distances on the validation set.
The independent critic is very generic.
The critic only needs samples from two distributions.

The approximate Wasserstein distance from the independent critic
allows us to compare different generator and critic architectures.
If we care about Wasserstein distances, we should be comparing generators
based on the approximate Wasserstein distance to the validation set.

The log-probability densities are less useful for generator comparison when the generators are generating only a subset of the data.
On the other hand, when doing lossless compression, 
we care about log-probabilities \citep{theis2015note}. 
When using real NVPs we can even jointly optimize both objectives (\fig{fig:mle_wgan_3} and \fig{fig:plot_obs_bits_mle_wgan}).

We show one additional usage of real NVPs for Adversarial Variational Bayes \citep{mescheder2017avae} evaluation in the appendix.

\begin{figure}[t]
\vskip 0.2in
\begin{center}
\centerline{\includegraphics[width=0.6\columnwidth]{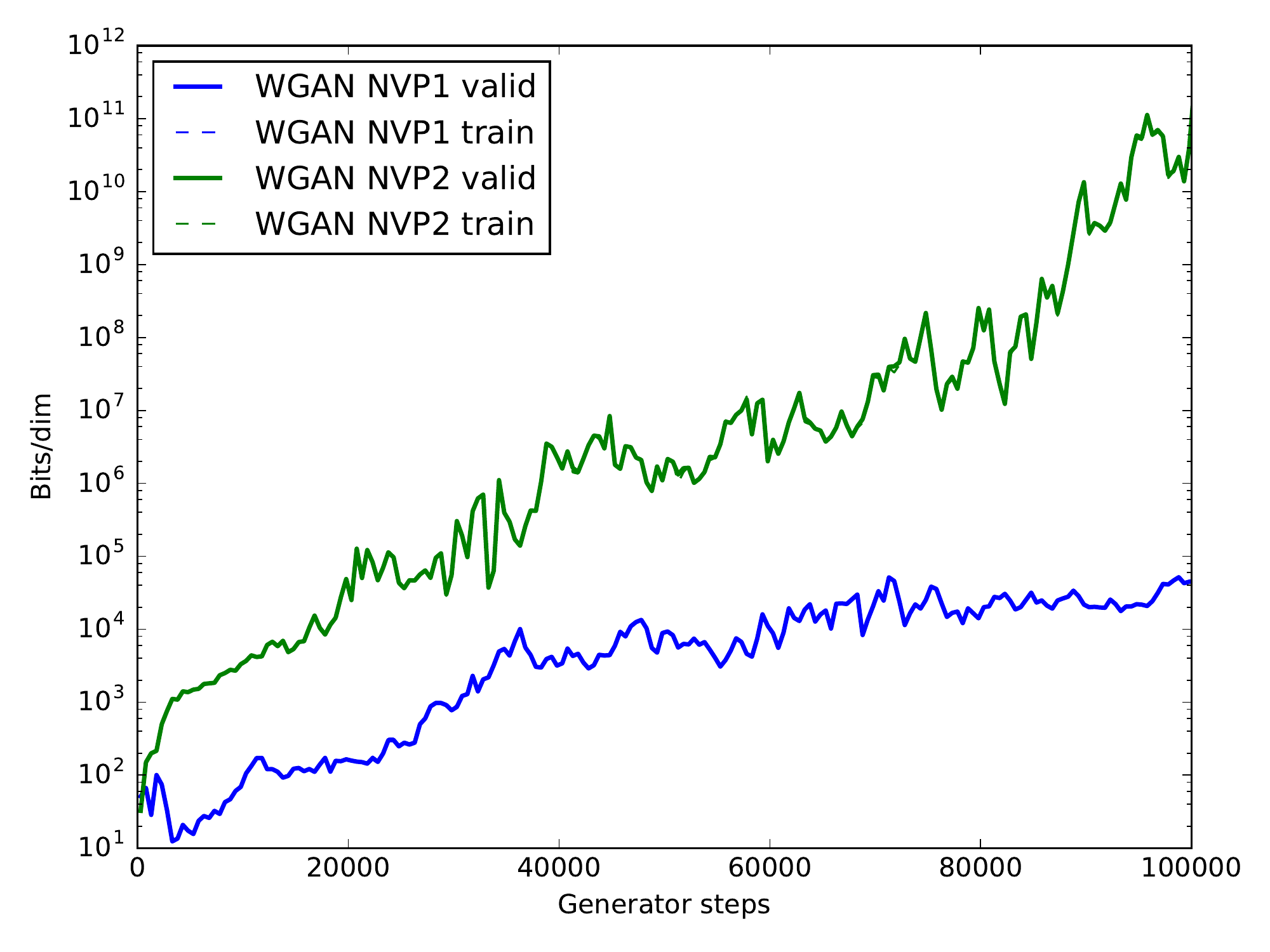}}
\caption{The negative log-probability density of real NVPs trained by WGAN on MNIST. The validation loss and the training loss overlap.}
\label{fig:plot_mnist_obs_bits_wgan}
\end{center}
\vskip -0.5in
\end{figure}

\begin{figure}[t]
\vskip 0.2in
\begin{center}
\begin{minipage}{0.23\textwidth}
\centerline{\includegraphics[width=\columnwidth]{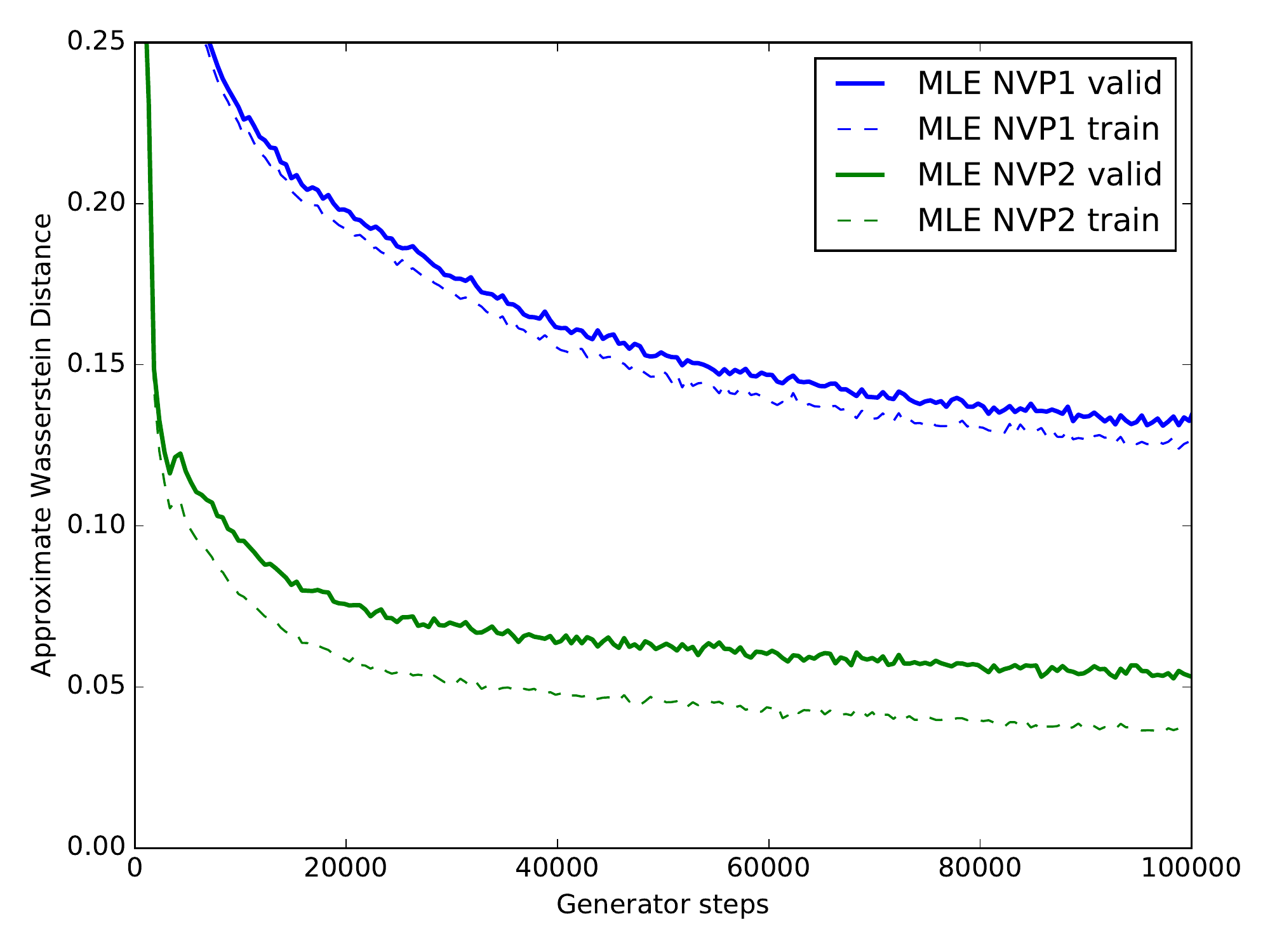}}
\end{minipage}
\begin{minipage}{0.23\textwidth}
\centerline{\includegraphics[width=\columnwidth]{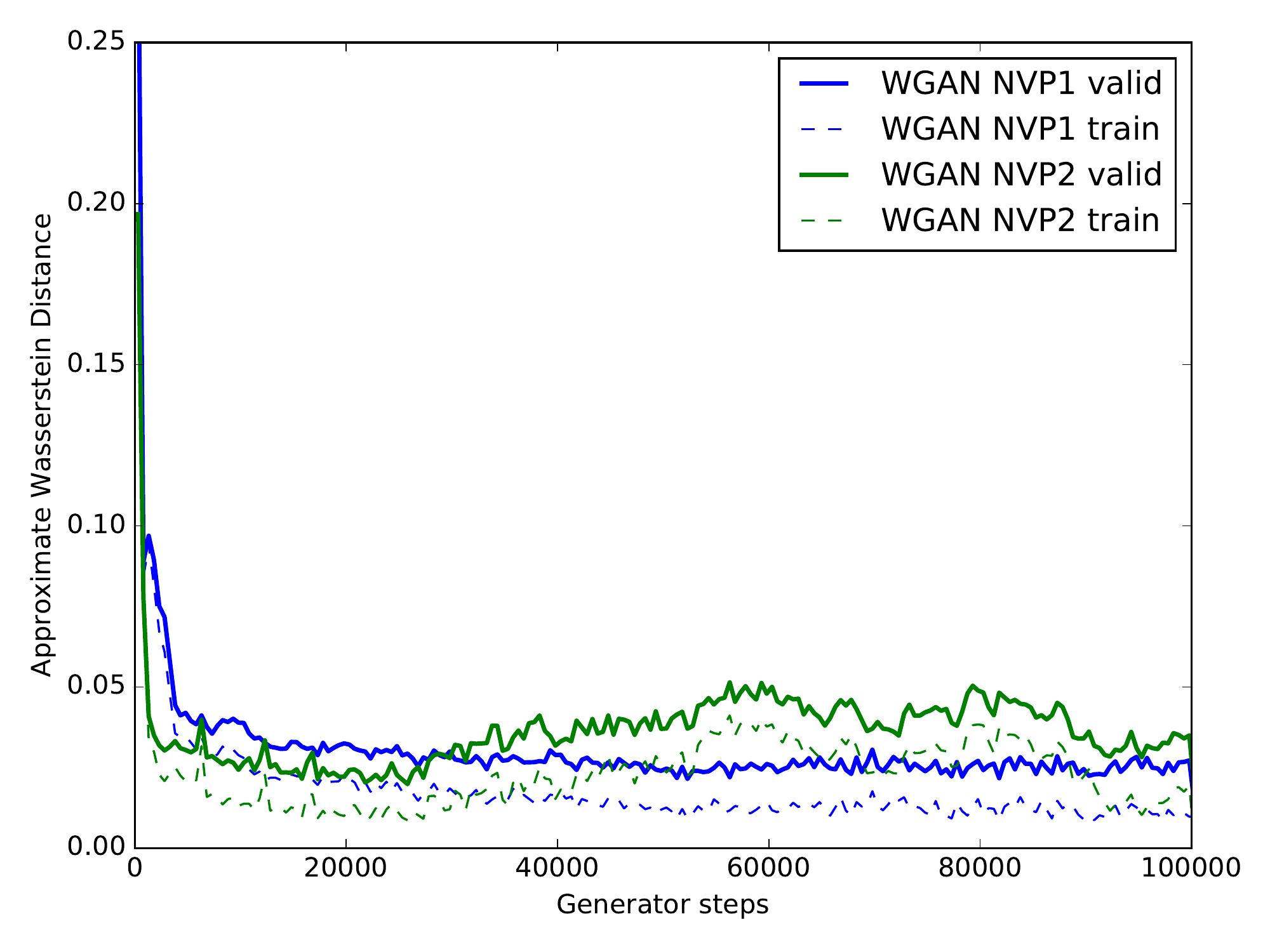}}
\end{minipage}
\caption{\textbf{Left:} The approximate Wasserstein distance for real NVPs trained by maximum likelihood on MNIST. \textbf{Right:} The approximate Wasserstein distance for real NVPs trained by WGAN. The Wasserstein distance was approximated by an independent critic.
}
\label{fig:plot_mnist_wdistance}
\end{center}
\vskip -0.2in
\end{figure} 

\section*{Acknowledgments}
We want to thank Mihaela Rosca, David Warde-Farley, 
Shakir Mohamed, Edward Lockhart, Arun Nair and Chris Burgess
for many inspiring discussions. We also thank the very helpful anonymous reviewer for suggesting to inspect the rank of the Jacobian matrix.

}{  

\appendix

\begin{figure}[t]
\vskip 0.2in
\begin{center}
\centerline{\includegraphics[width=0.8\columnwidth]{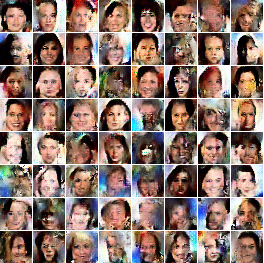}}
\caption{Samples generated from the shallow NVP1 trained by combined MLE+WGAN objectives. The samples are more globally coherent than samples from NVP1 trained by MLE alone (\fig{fig:nvp_3}-left).}
\label{fig:mle_wgan_3}
\end{center}
\vskip -0.2in
\end{figure} 

\begin{figure}[t]
\vskip 0.2in
\begin{center}
\centerline{\includegraphics[width=0.7\columnwidth]{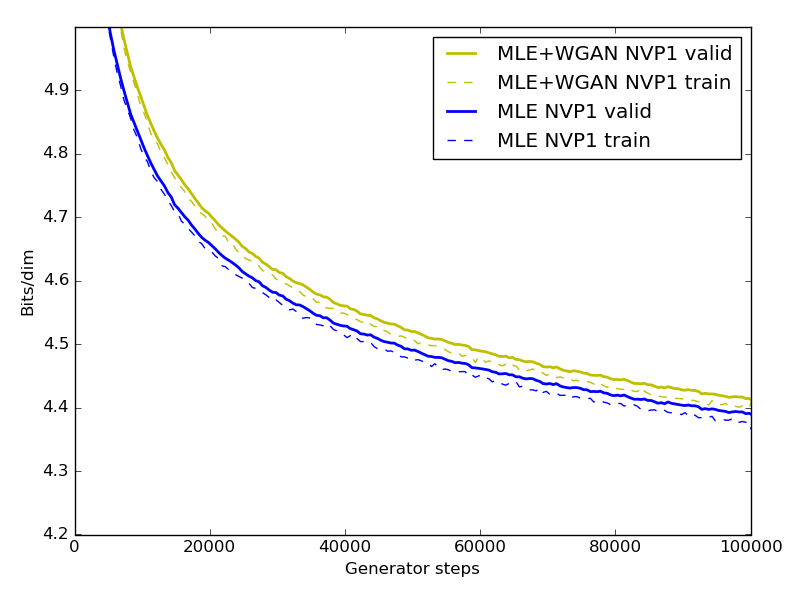}}
\caption{The negative log-probability density of NVP1 trained by combined MLE+WGAN objectives
is not much worse than the negative log-probability of NVP1 trained by MLE alone.}
\label{fig:plot_obs_bits_mle_wgan}
\end{center}
\vskip -0.2in
\end{figure} 

\begin{figure}[t]
\vskip 0.2in
\begin{center}
\centerline{\includegraphics[width=0.7\columnwidth]{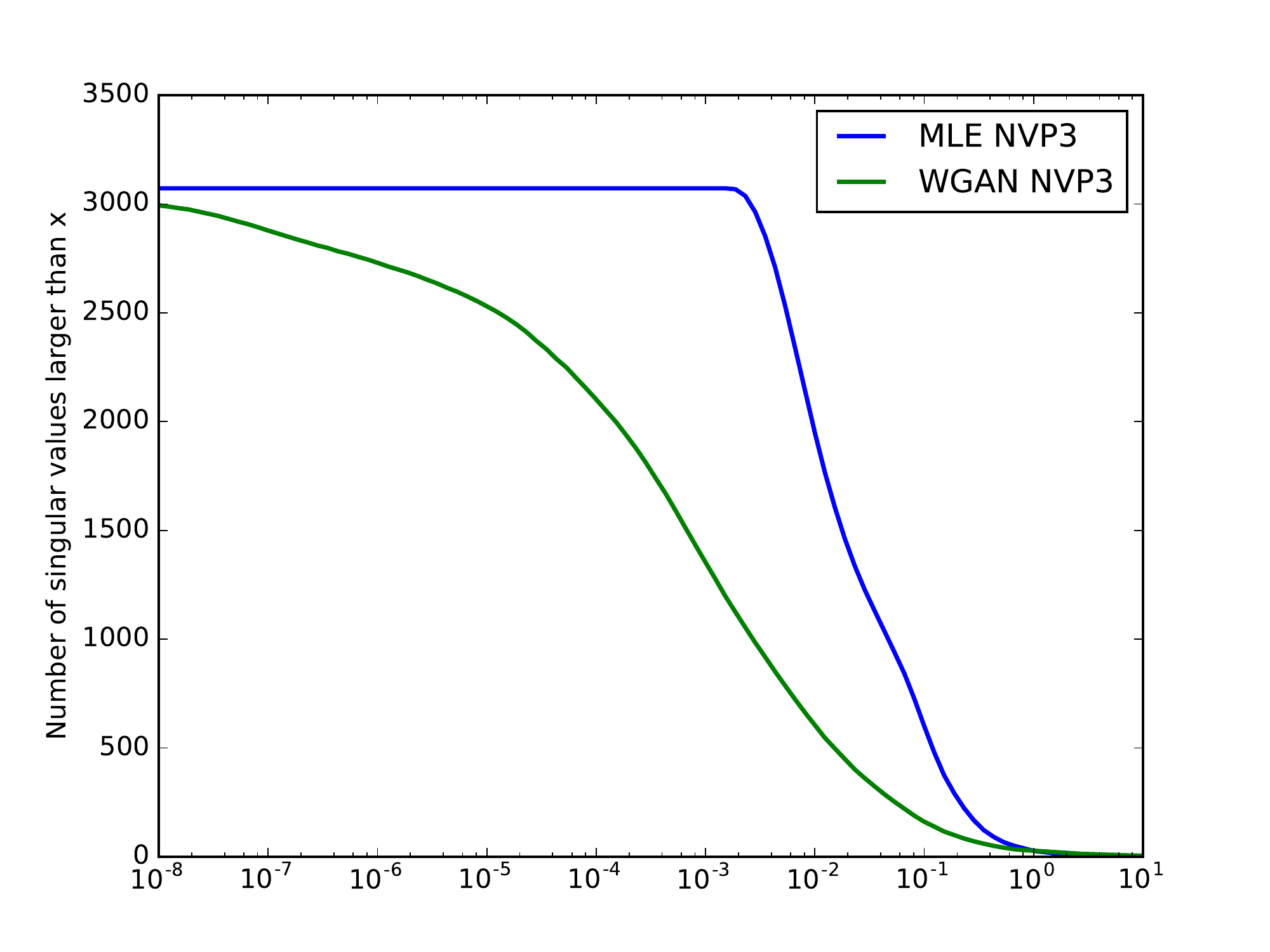}}
\caption{Inspection of the rank of the Jacobian matrix of the generator transformation for CelebA $32 \times 32$. The WGAN generator has a Jacobian with a low rank. The WGAN distribution then lies on a low dimensional manifold
as predicted by \citet{arjovsky2017towards}.}
\label{fig:plot_sval}
\end{center}
\vskip -0.2in
\end{figure}

\section{Log-probability Density Ratio Evaluation}
\label{sec:avae}

\begin{figure}[t]
\vskip 0.2in
\begin{center}
\centerline{\includegraphics[width=0.7\columnwidth]{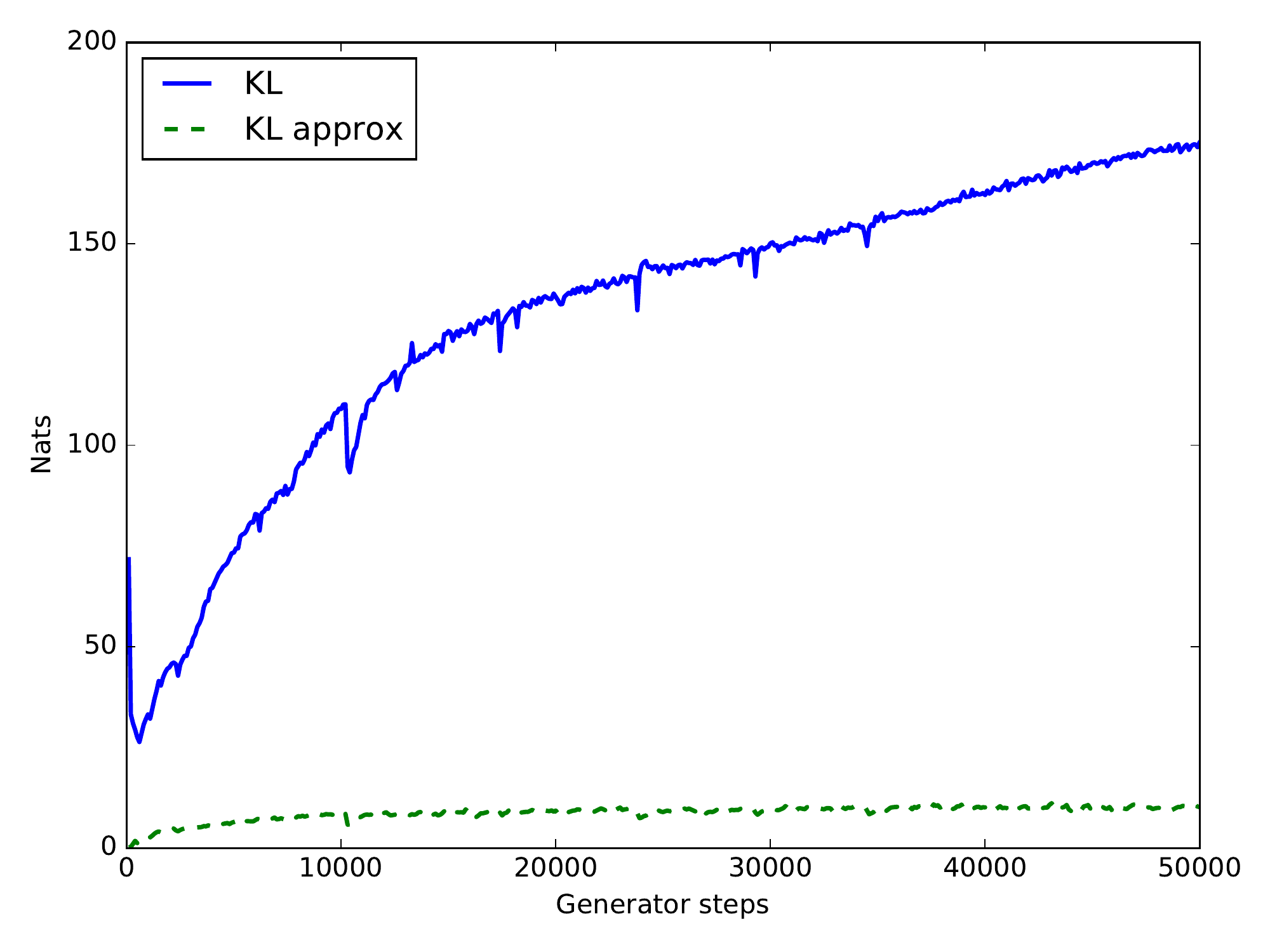}}
\caption{The unbiased $\kl{q(z|x)}{p(z)}$ value and the approximation of the $\klit$ from a discriminator. The discriminator was trained to recognize samples from the encoder from the samples from the prior. The discriminator is not representing well large $\klit$ divergences.}
\label{fig:plot_avae_kl}
\end{center}
\vskip -0.2in
\end{figure}

Real NVPs are not limited to the computation of negative log-probability densities of visible variables. For example, a real NVP can be used as an encoder in Adversarial Variational Bayes \citep{huszar2017avae,mescheder2017avae}. We were able to measure the gap between the unbiased $\klit$ estimate $\log q(z|x) - \log p(z)$ and its approximation from GAN. \fig{fig:plot_avae_kl} shows that Adversarial Variational Bayes underestimates the $\klit$ divergence. The discriminator would need to output $\logit(D(x))=-\klit$ to represent the $\klit$.

After measuring the problem, we can start thinking how to mitigate it. It would be possible to use auto-regressive discriminators \citep{oord2016conditional} to decompose the large $\klit$ divergence to multiple smaller terms:
\begin{align}
   \log \frac{q(z|x)}{p(z)} &= \sum_i \log q(z_i|x, z_{1:i-1}) - \log p(z_i)
\end{align}
where $p(z_i)$ is the independent Gaussian prior.

\clearpage
} 

\bibliography{main}
\bibliographystyle{icml2017}

\end{document}